%% file: neurips_2025.tex
\documentclass{article}

\PassOptionsToPackage{numbers, compress}{natbib}
\usepackage[preprint]{neurips_2025}
\usepackage{booktabs}       
\usepackage{threeparttable} 

\usepackage[utf8]{inputenc} 
\usepackage[T1]{fontenc}    
\usepackage{hyperref}       
\usepackage{url}            
\usepackage{booktabs}       
\usepackage{amsfonts}       
\usepackage{nicefrac}       
\usepackage{microtype}      
\usepackage{xcolor}         
\usepackage{graphicx}
\usepackage{amssymb}
\usepackage{multirow}
\usepackage[normalem]{ulem}
\useunder{\uline}{\ul}{}
\usepackage{makecell}
\usepackage{enumitem}
\usepackage{caption}
\usepackage{float}
\usepackage{fvextra}

\newtheorem{definition}{Definition}

\title{CrimeMind: Simulating Urban Crime with Multi-Modal LLM Agents}

\author{
 \textbf{Qingbin Zeng\textsuperscript{1}\thanks{Equal contribution.}},
 \textbf{Ruotong Zhao\textsuperscript{1}\footnotemark[1]},
 \textbf{Jinzhu Mao\textsuperscript{1}},
 \textbf{Haoyang Li\textsuperscript{2}},
 \textbf{Fengli Xu\textsuperscript{1}\thanks{Corresponding author: fenglixu@tsinghua.edu.cn}},
 \textbf{Yong Li\textsuperscript{1}}
    \\
    \\
 \textsuperscript{1}Department of Electronic Engineering, Tsinghua University\\
 \textsuperscript{2}Department of Sociology, Hong Kong Baptist University 
}

\begin{document}

\maketitle

\begin{abstract}

Modeling urban crime is an important yet challenging task that requires understanding the subtle visual, social, and cultural cues embedded in urban environments. Previous work has predominantly focused on rule-based agent-based modeling (ABM) and deep learning methods. ABMs offer interpretability of internal mechanisms but exhibit limited predictive accuracy. In contrast, deep learning methods are often effective in prediction but are less interpretable and require extensive training data. Moreover, both lines of work lack the cognitive flexibility to adapt to changing environments. 
Leveraging the capabilities of large language models (LLMs), we propose \textbf{CrimeMind}, a novel LLM-driven ABM framework for simulating urban crime within a multi-modal urban context. A key innovation of our design is the integration of the \textit{Routine Activity Theory} (RAT) into the agentic workflow of CrimeMind, enabling it to process rich multi-modal urban features and reason about criminal behavior. 
However, RAT requires LLM agents to infer subtle cues in evaluating environmental safety as part of assessing guardianship, which can be challenging for LLMs. 
To address this, we collect a small-scale human-annotated dataset and align CrimeMind’s perception with human judgment via a training-free textual gradient method.
More importantly, we leverage the commonsense reasoning capabilities of LLMs, which enable CrimeMind to conduct counterfactual simulations. Experiments across four major U.S. cities demonstrate that CrimeMind outperforms both traditional ABMs and deep learning baselines in crime hotspot prediction and spatial distribution accuracy, achieving up to a 24\% improvement over the strongest baseline. Furthermore, we conduct counterfactual simulations of external incidents (e.g., Black Lives Matter protests) and policy interventions (e.g., the Dallas Police Redistribution Plan). CrimeMind successfully captures the expected changes in crime patterns, demonstrating its ability to reflect counterfactual scenarios. Overall, CrimeMind enables fine-grained modeling of individual behaviors and facilitates evaluation of real-world interventions.
Our code is open-source at \url{https://anonymous.4open.science/r/CrimeMind-EB3E}.

\end{abstract}

\input{1_intro}
\input{2_related}

\input{3_problem}
\input{4_methods}
\input{5_experiments}

\input{6_conclusion}
\bibliographystyle{unsrt}
\bibliography{references}
\input{7_Appendix}

\end{document}

%% file: 1_intro.tex
\section{Introduction} 
\label{sec: intro}

Understanding and simulating urban crime is a central challenge at the intersection of computational science, criminology, and urban planning~\cite{groff2019state}. As cities grow in population, density, and inequality, crime increasingly reflects not only socioeconomic disparities but also complex interactions between individuals and their physical surroundings. Predicting where and why crimes happen requires models that account for subtle cues in the built environment—such as visual disorder, perceived safety, and social activity patterns, as well as the behaviors and intentions of individuals. These demands go far beyond simple statistical correlations or historical trend extrapolation, making urban crime modeling a uniquely difficult and socially impactful task.

Existing approaches generally fall into two categories: data-driven supervised learning and rule-based agent-based models (ABMs). Supervised models are often effective at forecasting crime hotspots~\cite{naik2014streetscore, rotaru2022event, wang2024diffcrime}, but they rely heavily on historical data, lack interpretability, and cannot simulate interventions or unseen scenarios. In contrast, ABMs offer interpretability and support counterfactual analysis~\cite{malleson2010crime, zhu2021agent}, but often rely on hand-crafted heuristics that exhibit limited predictive accuracy. Importantly, neither paradigm offers the cognitive flexibility needed to capture the situated, adaptive nature of criminal decision-making in complex and evolving urban contexts.

Recent advances in large language models (LLMs) provide a potential way to overcome these limitations. LLMs demonstrate strong capabilities in commonsense reasoning and role-playing~\cite{shanahan2023role}, and recent work has used LLM-based agents to simulate macroeconomic trends~\cite{li2023econagent}, cultural evolution~\cite{dai2024artificial}, and opinion polarization~\cite{piao2025emergence}. However, current LLM-based generative agent frameworks typically focus on general planning or social routines, and lack integration with domain-specific social theories. More importantly, they struggle to interpret multi-modal urban environments, which are critical for modeling real-world criminal behavior.

To bridge these gaps, we introduce \textbf{CrimeMind}, an LLM-driven agent-based framework for simulating urban crime in multimodal contexts. A key innovation of our approach is the integration of \textit{Routine Activity Theory} (RAT)~\cite{miro2014routine} into the cognitive architecture of LLM-powered agents, enabling theory-grounded and context-aware decision-making. 
Incorporating RAT into agent behavior allows for interpretable, theory-consistent reasoning about crime events. However, applying RAT in practice requires agents to infer subtle environmental cues to assess guardianship and situational risk—an ability that remains challenging for LLMs due to their limited perceptual grounding.
To address this, we collect a small-scale human-annotated dataset and employ a training-free textual gradient alignment method based on TextGrad~\cite{yuksekgonul2025optimizing}, aligning the visual perception of agents with human judgments. This alignment significantly improves the correlation between predicted and human-rated safety scores, increasing from 42\% to 79\%.
Criminal agents reason dynamically over RAT—identifying targets, estimating risks, and adapting behavior—guided by in-context LLM inference.
This design enables CrimeMind not only to replicate real-world crime distributions but also to simulate counterfactual scenarios by leveraging the commonsense reasoning capabilities of LLMs.

Extensive experiments across four major U.S. cities demonstrate that CrimeMind outperforms traditional agent-based models and deep learning baselines in both hotspot prediction and distributional accuracy, achieving over 24\% relative enhancement on average. Our ablation studies further reveal the critical contributions of theory-guided reasoning, multimodal urban context, and the general reasoning abilities of the LLM. Moreover, we conduct counterfactual simulations of external shocks (e.g., Black Lives Matter protests) and policy interventions (e.g., the Police Redistribution Plan in Dallas). CrimeMind successfully captures expected changes in crime patterns, demonstrating its ability to reflect counterfactual scenarios, enabling “what-if” analyses for urban safety planning.

In summary, the main contributions of this work include:

\begin{itemize}[leftmargin=15pt]
    \item We propose the first generative agent-based crime simulation framework that explicitly integrates criminological theory into cognitive architecture of LLM agents, enabling interpretable and realistic modeling of criminal behavior in complex urban environments. 
    \item We calibrate agents’ perception of visual safety with human judgments via a training-free textual gradiant approach. This alignment allows agents to infer subtle environmental cues necessary for reproducing more human-like crime patterns.
    \item Through extensive experiments, we demonstrate that CrimeMind outperforms existing baselines in crime simulation, and more importantly, it enables counterfactual simulations under social shocks and policy interventions, offering a valuable tool for urban safety evaluation and planning. 
\end{itemize}

%% file: 2_related.tex
\section{Related Work}
\paragraph{Crime Simulation}
Recent studies have shown that understanding and predicting urban crime is a significant social and scientific challenge~\cite{groff2019state, jenga2023machine}. 
Deep learning approaches, such as the Granger Network~\cite{rotaru2022event} and DiffCrime~\cite{wang2024diffcrime}, have demonstrated strong performance, but the black-box nature limits the interpretability and counterfactual reasoning. In contrast, traditional ABMs simulate criminal behavior by encoding decision-making rules derived from criminological theories such as Routine Activity Theory~\cite{miro2014routine} or Rational Choice Theory~\cite{scott2000rational}. These models provide more interpretable simulations and applications in burglary prevention~\cite{malleson2010crime} and police strategy~\cite{weisburd2017can, zhu2021agent}. However, these models often oversimplify agent cognition, ignoring bounded rationality, perceptual cues, and agent heterogeneity. To address these limitations, we propose an LLM-driven agent-based framework that is grounded in criminological theory and enriched with urban context. This approach enhances both the realism and the interpretability of simulated crimes by capturing complex, context-aware dynamics.


\paragraph{Generative Agents in Social Simulation}

With the emergent role-play and reasoning capabilities of LLMs~\cite{xu2025towards}, generative agents have emerged as powerful tools for simulating human behavior and social dynamics~\cite{shao2024chain, park2024generative, gurcan2024llmaugmented,hewitt2024predicting}. Recent studies demonstrate their potential to model complex social phenomena such as macroeconomics~\cite{li2023econagent}, social evolution~\cite{dai2024artificial}, segregation~\cite{fan2025invisible, yan2024opencity}, and opinion polarization~\cite{piao2025emergence}. In large-scale social simulation, benchmark challenges such as AgentSociety~\cite{yan2025agentsociety} further stimulate research on designing capable, personalised LLM agents, while work on meta-structure discovery in heterogeneous information networks~\cite{chen2024large} highlights the versatility of LLM reasoning across complex relational contexts. 
These methods further show that generative agents perform well in social experiments utilizing demographic data~\cite{hewitt2024predicting}. 
Although generative agents have shown promise across various social science domains, their direct application to crime simulation remains limited. Existing models typically lack grounding in criminological theory and fail to reproduce criminal behavior. In contrast, our work introduces a theory-driven agent design grounded in Routine Activity Theory, which ensures that agents' decisions reflect criminological realism, supporting more accurate and interpretable simulations of urban crime patterns.



\paragraph{Multi-modal Urban Computing}
The strong relationship between multi-modal information and urban crime patterns makes visual urban computing a critical research direction in crime prediction. 
In recent years, visual urban computing has been widely used in socioeconomic estimation \cite{li2022predicting}, urban vitality analysis \cite{shi2019visualization}, and safety risk forecasting \cite{zhang2020multi} through the analysis of street-view images, remote sensing imagery, and semantic urban data. For example, recent work links human mobility traces to experienced inequalities across cities~\cite{xu2025using}, while Naik et al.\ employed street-view imagery to estimate income levels and perceived safety~\cite{naik2014streetscore, naik2017computer}. 
However, most of these approaches rely on statistical correlations between visual features and socio-economic indicators, often lacking explanatory mechanisms for human behavior~\cite{mandalapu2023crime}. In contrast, recent advances demonstrate that LLM-based agents are capable of perceiving fine-grained spatial cues from urban imagery and reasoning within cognitively grounded frameworks~\cite{zeng2024perceive}. Building on this, our proposed CrimeMind framework embed visual and semantic features into LLM-driven agents' decision-making process, enabling fine-grained simulations of individual criminal behavior in visually grounded environments.


%% file: 3_problem.tex
\section{Problem Formulation}
\label{sec: problem}

\subsection{Agent-based crime simulation}
To simulate crime dynamics in urban environments, we define an agent-based crime simulation framework in which heterogeneous agents interact with structured spatial environments through mobility and decision-making processes.

Let the urban environment be denoted as $E$, spatially discretized into a set of grid-level regions $\mathcal{G} = \{g_1, g_2, \ldots, g_N\}$, such as Census Block Groups (CBGs). Each grid cell $g \in \mathcal{G}$ is associated with a set of static features $\phi(g)$, including point-of-interest (POI) types, demographic indicators, economic statistics, and visual features (e.g., street view safety scores). The agent population is defined as $A = \{R, C, P\}$, where $R$ is the set of ordinary residents, $C$ is the set of potential criminals, and $P$ is the set of police agents. At each discrete time step $t$, each agent $a \in A$ occupies a grid cell $g_a^t \in \mathcal{G}$. All agents make mobility decisions based on spatial features and personal preferences. Specifically, resident agents $r \in R$ and criminal agents $c \in C$ move across the grid to simulate daily routines, governed by empirically validated mobility models such as the EPR model~\cite{song2010modelling}. And police agents $p \in P$ patrol within their jurisdictional areas, contributing to guardianship intensity. 

Criminal decision-making is the focus of our work. At each step $t$, a potential criminal agent $c \in C$ assesses whether to commit a crime based on their own state (e.g., prior behavior, motivation), the presence and characteristics of nearby agents (potential victims and police), and the local environment $\phi(g_c^t)$, where $g_c^t$ is the agent’s current grid. Formally, the crime decision can be modeled as a function:
\begin{equation}
\mathcal{D}_c^t = f_\theta(\phi(g_c^t), \mathcal{N}_c^t, h_c^t),
\end{equation}
where $\mathcal{D}_c^t \in \{0, 1\}$ is the binary crime decision of agent $c$ at time $t$; $\mathcal{N}_c^t$ represents the set of neighboring agents within vicinity; $h_c^t$ denotes the internal state (e.g., memory or motivation) of the agent; $f_\theta$ is a policy function driven by predefined rules or LLM reasoning.

Over a full simulation period of $T$ steps, the model outputs a sequence of crime events:
\begin{equation}
\mathcal{C} = \bigcup_{t=1}^T \{(c, g_c^t, t) \mid \mathcal{D}_c^t = 1 \},
\end{equation}
which can be aggregated to construct spatiotemporal crime heatmaps for evaluation or policy testing. A simple illustration of agent-based crime simulation is shown in Figure \ref{Fig:hotspots}a.

\subsection{Datasets and statistics}
\label{subsec: dataset}
To construct a realistic urban simulation environment, initialize agent profiles, and evaluate model outputs, we leverage multiple heterogeneous datasets, including demographic statistics, street-level imagery, and historical crime records.

\textbf{Demographic and Environmental Features.}  
We utilize Census Block Group (CBG)-level demographic data from SafeGraph, which includes variables such as median household income, racial composition, gender ratios, and population density. These features serve dual purposes: (1) they form part of the static environment representation $\phi(g)$ for each CBG $g \in \mathcal{G}$; (2) they are sampled to initialize the profile distributions of resident and criminal agents, enabling agent-level heterogeneity. All data used in this work correspond to the year 2019, ensuring temporal consistency across modalities.

\textbf{Street View Imagery and Visual Perception.} 
Visual features of the environment are extracted from Google Street View (GSV) images, collected via the Google Street View API around 2019. For each CBG, we retrieve multiple panoramic images at street-level viewpoints. These images are processed by a vision-language model (VLM) to extract two types of perceptual cues: \textit{perceived safety score}, quantifying the subjective visual safety of a location; and \textit{semantic description }of the scene, which serves as input to the LLM-based agent for contextual awareness. To ensure robust visual representation, we aggregate safety scores and semantic outputs across all images within a CBG. CBGs with insufficient street view coverage are excluded from the simulation.

\begin{figure}[t]
    \begin{center}
        \includegraphics[width=1.0\linewidth]{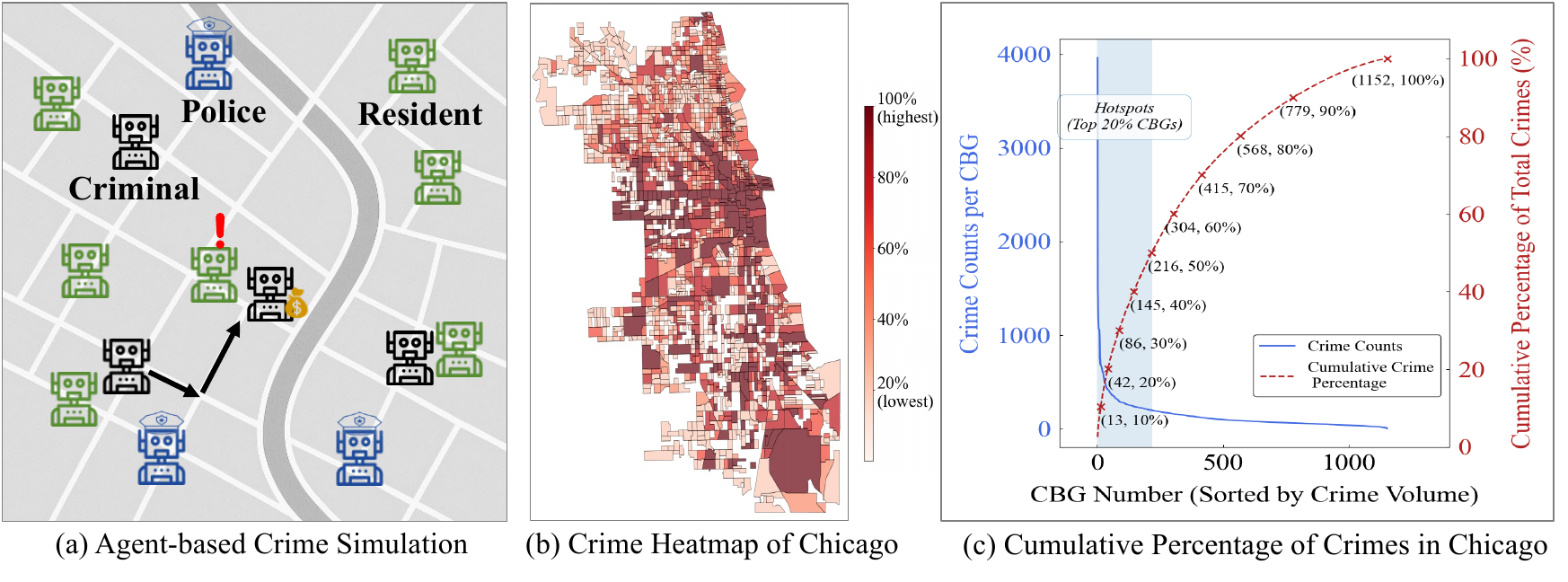}
        \caption{Illustration of problem and hotspot definitions}   
        \label{Fig:hotspots}
    \end{center}
\vspace{-0.5cm}
    
\end{figure}

\textbf{Crime Records and Hotspot Identification.} 
Ground-truth crime data is sourced from official open-data portals. For example, the data of Chicago is from the official data portal\footnote{https://data.cityofchicago.org/}, which includes timestamped incident reports with geographic coordinates and crime categories. To enable evaluation and visualization, we aggregate crime events over a given period to construct spatial crime distributions. As illustrated in Figure \ref{Fig:hotspots}b (spatial crime distribution) and Figure \ref{Fig:hotspots}c (cumulative crime coverage), crime incidents are highly spatially concentrated. In Chicago, approximately 50\% of crimes occur within just 20\% of the CBGs, indicating the existence of distinct crime hotspots.
We formally define the crime hotspots as follows:
\begin{definition}[Crime Hotspots]
    Let $\mathcal{C} = \{(g_i, t_i)\}$ denote the set of crime events, where each crime occurs at grid $g_i \in \mathcal{G}$ and time $t_i$. For a fixed time window, we compute the total crime count $c(g)$ for each grid $g$. Sorting $\{c(g)\}$ in descending order, we define the top $\alpha\%$ of grids that cumulatively account for $\beta\%$ of all crimes as crime hotspots $\mathcal{H}$, with typical values $\alpha = 20\%$, $\beta = 50\%$.
\end{definition}
This definition provides an empirical and replicable standard for hotspot detection, serving as the ground truth for evaluating our simulation’s predictive accuracy. In this work, we define the top $\alpha=20\%$ CBG in ground truth as crime hotspots.

\subsection{Evaluation Protocol}
\label{subsec: evaluation}
The central research problem of this paper is to model and analyze the spatiotemporal dynamics of urban crime using an agent-based simulation framework powered by LLMs. The goal is to simulate realistic crime events $\mathcal{C}_{\mathrm{sim}}$, and evaluate how well they replicate real-world crime patterns $\mathcal{C}_{\mathrm{real}}$ across both space and time. Given the simulation process produces a sequence of crime events $\mathcal{C}_{\mathrm{sim}} = \{(t_i, g_i)\}$, where each event is associated with a grid (CBG) $g_i$ and time $t_i$. These events are aggregated at the grid level and normalized to mitigate biases from absolute crime volume.

We evaluate the model using three quantitative metrics: Jensen-Shannon Divergence (JSD), Root Mean Squared Error (RMSE), and Crime Hotspot Hit Rate (HR@K). JSD and RMSE are used to assess the overall similarity between the real and simulated crime distribution in terms of divergence and absolute error. The crime hotspot hit rate is a newly defined metric inspired by top-$K$ accuracy in the recommendation system, and we define it formally as follows:

\begin{definition}[Crime Hotspot Hit Rate(HR@K)]
    Let $\mathcal{G}$ be the set of all grid cells, and define 
    $\mathcal{H}_{\mathrm{real}} \subset \mathcal{G}$: the set of real crime hotspots, defined as the top grids that account for 50\% of total observed crimes (typically the top 20\% of CBGs);
    $\mathcal{H}_{\mathrm{sim}}^{(K)} \subset \mathcal{G}$: the top $K \times |\mathcal{H}_{\mathrm{real}}|$ grids ranked by simulated crime count.
    Then the Hit Rate can be defined as:
    \begin{equation}
        \mathrm{HR@K} = \frac{|\mathcal{H}_{\mathrm{real}} \cap \mathcal{H}_{\mathrm{sim}}^{(K)}|}{|\mathcal{H}_{\mathrm{real}}|}
    \end{equation}
\end{definition}

We report results under multiple values of $K$ (e.g., 1.0, 1.5,2.0) to assess the robustness of hotspot prediction under varying recall thresholds. A higher HR@K indicates stronger alignment between the simulated and empirical spatial concentration of crimes.
These metrics allow us to systematically evaluate the model’s ability to capture both broad distributional patterns and localized high-risk areas—crucial for practical applications such as hotspot policing and urban planning.


%% file: 4_methods.tex
\section{CrimeMind Framework}
\label{sec: methods}
\subsection{Routine Activity Theory-Guided Agentic Architecture}
\begin{figure}[ht]
    \begin{center}
        \includegraphics[width=1.0\linewidth]{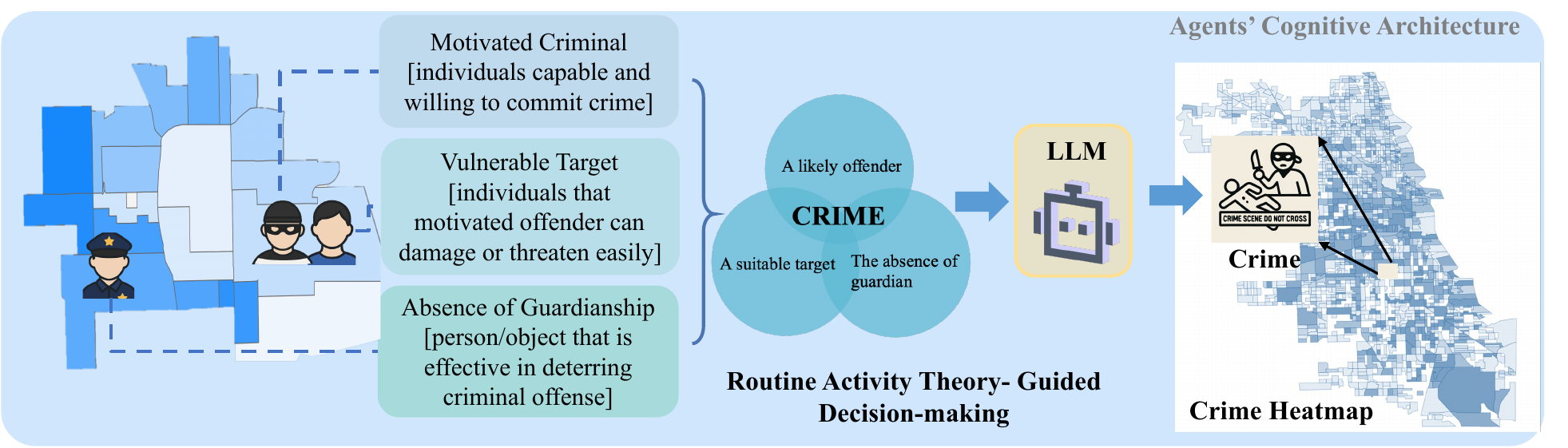}
        \caption{Routine Activity Theory-Guided Agentic Architecture}   
        \label{Fig:design}
    \end{center}
\end{figure}

Our framework, CrimeMind, leverages theory-grounded LLM agents embedded in an agent-based simulation to model and predict urban crime. Inspired by the Routine Activity Theory (RAT~\cite{miro2014routine}, we design LLM-powered agents whose decision-making is guided by three core components: (1) the motivation of the offender, (2) the vulnerability of the target, and (3) the absence of capable guardianship. These elements are encoded into a structured reasoning pipeline that enables transparent, interpretable, and context-aware crime simulations. Each agent (citizen, criminal, or police) operates within a grid-based urban environment, where each cell corresponds to a Census Block Group. Agent mobility follows an Exploration and Preferential Return (EPR) process~\cite{song2010modelling}, approximating realistic human movement across urban areas. At each simulation step, a criminal agent evaluates its intent to commit a crime based on the RAT-aligned cognitive architecture.

Specifically, \textbf{motivation} is derived from an agent's static profile (e.g., socio-economic attributes) and dynamic behavioral history (e.g., prior successes or failures in committing crimes). \textbf{Target vulnerability} is assessed when encountering citizen agents, factoring in demographic attributes and multimodal environmental information—such as census data and street-level imagery. Visual cues are processed by a Vision-Language Model (VLM), which generates semantic summaries of environment. \textbf{Guardianship} is captured by the presence and proximity of nearby police agents, as well as the perceived safety of the surrounding area, quantified by a VLM-predicted visual safety score.

These RAT components are fused into a structured prompt that is passed to the LLM-based criminal agent. The LLM returns both a binary crime decision and a natural language justification, which enables interpretability and supports counterfactual interventions (e.g., increasing patrol density or altering the visual appearance of a street). This architecture, as illustrated in Figure~\ref{Fig:design}, is modular and extensible, and, to our knowledge, represents the first integration of RAT into the cognitive reasoning pipeline of LLM-based generative agents for crime simulation.

\begin{figure}[htb]
    \begin{center}
        \includegraphics[width=1.0\linewidth]{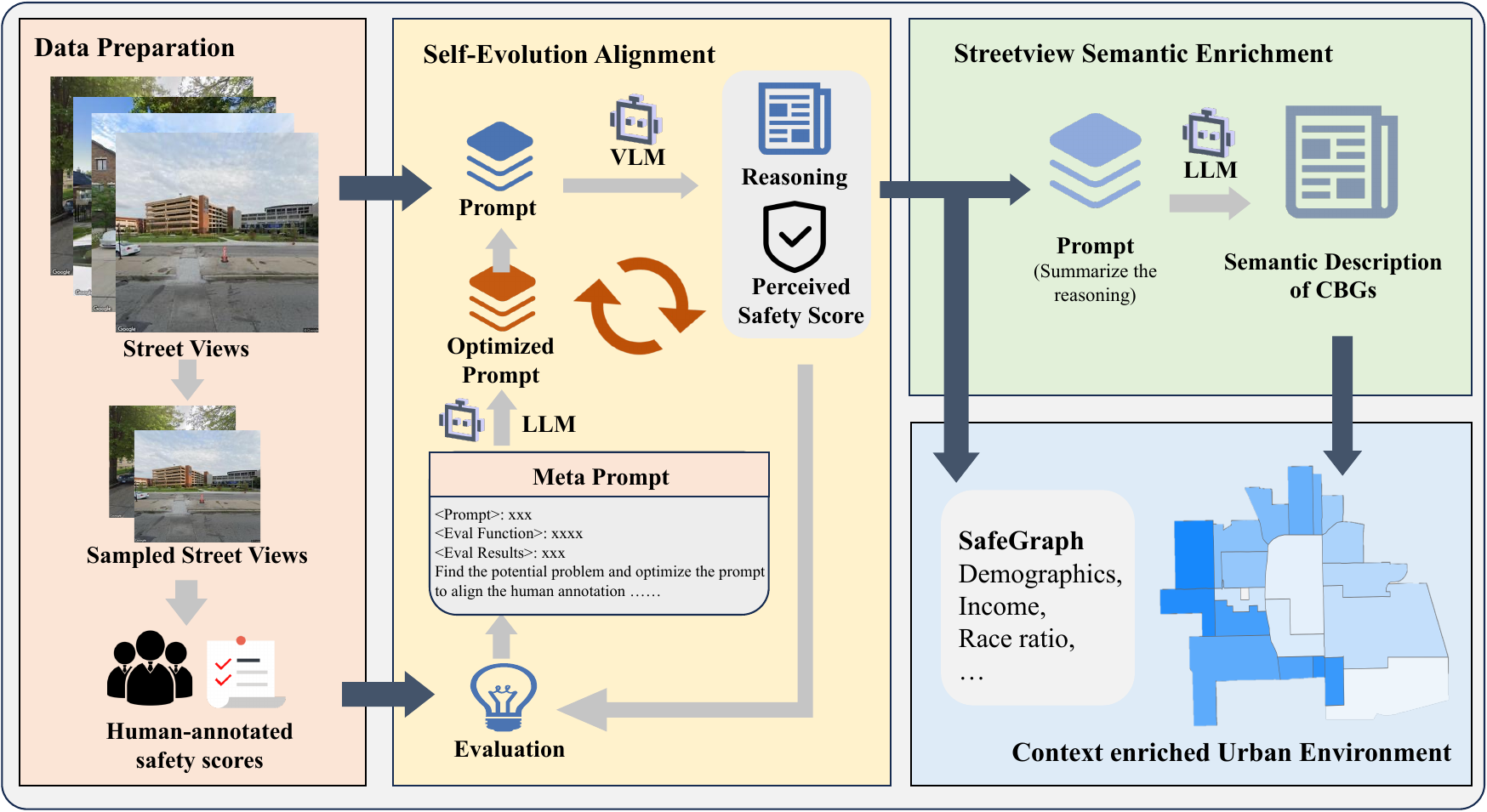}
        \caption{Knowledge Guided Agentic Workflow for Urban Environment Construction}   
        \label{Fig:pipeline}
    \end{center}
\vspace{-0.3cm}
\end{figure}

\subsection{Mutli-modal Urban Cues Perception}
To enable interpretable and theory-informed decision-making within LLM agents, we construct a urban environment that integrates multi-modal urban features perception at the CBG level. This design aligns with Routine Activity Theory (RAT), which emphasizes the importance of environmental cues and socio-spatial context in shaping crime behavior. 

We begin by collecting structured data from SafeGraph and the U.S. Census, encompassing features such as population demographics, POI density, and average income. These attributes capture key aspects of neighborhood vulnerability and social structure, directly informing the RAT components of suitable targets and offender motivation.

Moreover, RAT highlights the role of perceived guardianship—often conveyed through the visual character of a place. To model this perceptual aspect, we collect Google Street View (GSV) images across each CBG and apply a VLM (Qwen2.5-VL-72B) to estimate perceived safety. Using carefully designed prompts, the model produces both numerical safety scores and qualitative descriptors for each image. To move beyond isolated visual cues toward a more comprehensive environmental understanding, we aggregate these outputs with a larger language model (Qwen2.5-72B), which generates high-level textual summaries for each CBG. These summaries reflect collective impressions—such as “well-maintained,” “dim lighting,” or “visible neglect”—that capture ambient perceptions of safety and aesthetic quality, influencing both criminal and civilian decision-making.

By integrating structured socio-demographic features with these perceptual and semantic cues, we construct a rich, multimodal representation of environment. This enables CrimeMind to reason over both factual and experiential aspects of space, producing more realistic and interpretable behavior.

\subsection{Human Judgment Alignment}
Perceptions of environmental safety are inherently subjective and cognitively complex. However, such perceptions play a critical role in shaping the decision-making processes guided by RAT. To ensure that our agents' understanding of visual environments reflects human intuition, we adopt a training-free alignment method that calibrates the outputs of a vision-language model (VLM) to human judgments. Figure~\ref{Fig:pipeline} illustrates the complete knowledge-guided alignment workflow.

\textbf{Human annotation.} We begin by collecting GSV images sampled from across each CBG. A subset of these images is then evaluated by human annotators, who rate the perceived safety using a three-point ordinal scale. To enhance consistency and reduce individual bias, we employ a comparative annotation strategy: annotators are shown triplets of images and asked to rank them from most to least safe. The final scores are aggregated by averaging across multiple comparisons, resulting in a high-quality, human-grounded reference dataset.

\textbf{Self-evolution alignment workflow.} We begin with a moderate alignment between VLM-predicted safety scores and human ratings (Pearson $r=0.42$). To improve this alignment, we implement a self-evolving prompt refinement process. A stronger LLM (GPT-4o) is provided with the current prompt, sample outputs, and alignment metrics, and iteratively proposes revised prompts and scoring criteria to better reflect human perception. This iterative process continues until convergence is achieved, at which point the optimized prompt is used to rescore all images citywide.

This alignment process ensures that agents' perception of environmental safety is not only grounded in visual input, but also reflective of human judgment, making the crime simulation more realism.

%% file: 5_experiments.tex
\section{Experiments}
\label{sec: exp}
\subsection{Experimental Setup}
\label{subsec: exp_setup}

We conduct extensive experiments to evaluate the effectiveness of our proposed framework and its capacity to generate actionable insights. 
The evaluation metrics are defined in Section \ref{subsec: evaluation}.
The simulation environment consists of thousands of CBGs across four cities (Chicago, New York, Dallas, and Los Angeles), each enriched with demographic, socioeconomic, and visual features. Agents include 4,000 citizens, 1,000 criminals (aligned with empirical findings suggesting approximately 20\% criminal propensity in populations\cite{archer1996criminal}), and 500 police officers, proportionally distributed across the CBGs. The simulation runs for 50 time steps, producing spatiotemporal sequences of crime events, aggregated at the CBG level for subsequent analysis.
\input{tables/main_table}

\subsection{Overall Performance}
We begin by evaluating CrimeMind on the city of Chicago, using Qwen2.5-7B-Instruct as the decision-making engine. We compare against a diverse set of baselines. The \textit{Random} baseline represents the lower bound with uniformly distributed decisions. The \textit{ABM series} includes representative agent-based crime models that follow rule-driven behaviors\cite{malleson2010crime, weisburd2017can, zhu2021agent}. \textit{UVI} is a vision-based method that leverages segmented street view features to regress CBG-level crime rates~\cite{fan2023urban}.

As shown in Table~\ref{tab:main_results}, CrimeMind outperforms all baselines across all metrics. In Chicago, it achieves the highest HR@2.0 (0.7157), indicating stronger top-k hotspot alignment, and yields the lowest JSD (0.0838) and RMSE (1.62E-04), reflecting accurate crime distribution modeling at the CBG level. Among traditional ABMs, ABM-Burglary shows relatively better HR@2.0 (0.6607) but suffers from higher JSD, revealing less stable spatial modeling. The vision-based UVI model underperforms in both spatial and numerical metrics, highlighting the limitations of static perception-based estimation in complex urban crime dynamics.
To examine generalizability, we extend our experiments to three additional U.S. cities: New York, Dallas, and Los Angeles. CrimeMind consistently achieves strong performance in all cities, maintaining high hotspot hit rates and low spatial divergence. For instance, in Dallas, CrimeMind achieves HR@2.0 of 0.6375 and JSD of 0.0887, outperforming all other baselines. Even in structurally diverse cities like New York, the framework achieves competitive results (HR@2.0 = 0.5164, JSD = 0.1750), validating its robustness.

These results collectively demonstrate that CrimeMind not only excels in simulating fine-grained spatial crime patterns in a single city but also generalizes effectively across different urban environments. Its ability to integrate multimodal context with LLM-based reasoning offers a significant advancement over traditional rule-based or perception-based methods in crime modeling. The ablation study in Appendix \ref{appendix: ablation} demonstrates that each module contributes to CrimeMind’s success. The combination of decision-theoretic modeling (via RAT), multi-modal urban context, and LLM reasoning provides a comprehensive foundation for accurate spatial crime simulation.
\vspace{-0.3cm}

\subsection{Evaluation of Perceived Safety Alignment}
We evaluate the alignment between model-predicted safety scores and human perception using a dataset of 100 street view images, each rated by ten annotators on a three-point safety scale. The averaged scores are normalized and discretized into 20 safety levels. Human ratings exhibit high internal consistency (Cronbach’s alpha = 99.26\%). The dataset is split 7:3 for training and evaluation.
Initial predictions from the VLM using a handcrafted prompt yield a Pearson correlation of 0.42 with human labels. Through iterative prompt optimization, guided by GPT-4o, using alignment metrics and scoring samples, the correlation improves substantially to 0.79 after several rounds, as shown in Appendix Figure \ref{Fig:textgrad}, demonstrating enhanced alignment with human perception.

After convergence, we apply the optimized prompt citywide to estimate perceived safety for all images. For each CBG, we compute the average scores and generate corresponding semantic summaries. These outputs serve as contextual inputs in the simulation. As shown in Table \ref{tab:ablation} in our ablation study, removing perceived safety signals leads to a substantial drop in HR@2.0 and increases in JSD and RMSE, underscoring their importance for modeling realistic spatial crime patterns.

\subsection{Counterfactual Analysis}
To see whether CrimeMind can reflect realistic shifts in crime distribution given new socio-political contexts, we conducted simulations in two representative scenarios: (i) external shocks (e.g., BLM protests) and (ii) policy interventions (e.g., Dallas Violent Crime Reduction Plan, 2020).

\textbf{External Shocks: Black Lives Matter (BLM) Protests.}
We simulate crime dynamics under external shocks that disrupt routine urban patterns, using the 2020 Black Lives Matter (BLM) protests in Chicago as a case study. Real-world data show a notable surge in violent crime during the protest period (Figure~\ref{Fig:appendix_crimeTime}). To replicate this, we introduce counterfactual conditions into the LLM agents' cognitive prompts, which describe the August 2020 BLM protest scenario, highlighting aspects such as heightened social tensions and protest activities. These contextual modifications aim to assess whether LLM-driven agents adapt their decisions under such socio-environmental disruptions. We evaluate the simulation against real crime data from August 2020, with a focus on whether the model captures known shifts, such as the emergence of temporary crime hotspots.

\begin{figure}[!t]
    \begin{center}
        \includegraphics[width=0.95\linewidth]{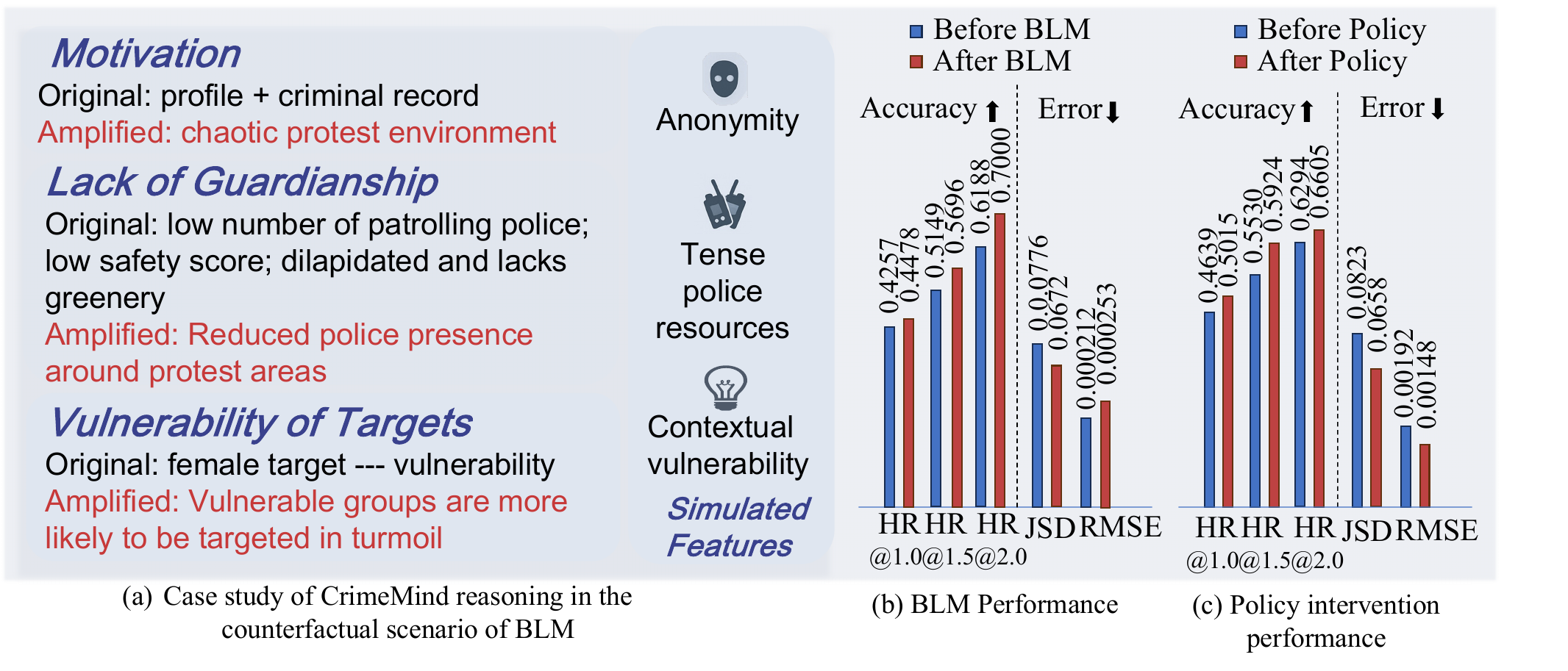}
        \caption{The counterfactual results of BLM and Policy intervention.}   
        \label{Fig:counterfactual}
    \end{center}
    \vspace{-0.3cm}
\end{figure}

Figure~\ref{Fig:counterfactual}a illustrates how the BLM context influences the core components of decision-making under RAT. As shown in Figure~\ref{Fig:counterfactual}b, incorporating BLM context improves hotspot prediction accuracy (HR@1.0 increases from 0.4257 to 0.4478) and spatial distribution fidelity (JSD decreases from 0.0776 to 0.0672). Most notably, the model more accurately identifies new, short-term hotspots observed during the protests (New Hotspot Concordance increases from 0.4902 to 0.5490).

\textbf{Policy Intervention: Police Redistribution Plan in Dallas.}
We further examine CrimeMind’s use for policy evaluation by simulating the effects of the “Violent Crime Reduction Plan” introduced by Dallas Mayor Eric Johnson in 2020. This real-world initiative prioritized reallocating police presence to high-crime areas and targeting repeat offenders. In our simulation, we operationalized this by: (1) implementing dynamic "hot spots policing," where police agents' patrolling logic was altered from random routes to assignments based on simulated real-time crime counts, thereby increasing police presence in higher-crime CBGs; and (2) incorporating a mechanism for offender removal, where the top 10 most active criminal agents were arrested at each step in the simulation.

Results in Figure~\ref{Fig:counterfactual}c show a substantial reduction in overall crime (from 7926 to 5218 incidents), and a sharper decline in hotspot crime (from 4113 to 2101), leading to a reduced hotspot crime ratio (from 0.5189 to 0.4026). Furthermore, when evaluating the model's hotspot predictions against actual crime data from Dallas for the 2021-2024 period (representing a post-policy implementation timeframe), the HR@1.0 improved from 0.4639 to 0.5015. Other distributional metrics also showed positive changes, with JSD decreasing from 0.0823 to 0.0658 and RMSE falling from 0.00192 to 0.00148. These improvements demonstrate that CrimeMind is capable of simulating the reasonably long-term effects of policies that involve strategic resource deployment.

These experiments highlight a key contribution of CrimeMind: it facilitates counterfactual reasoning and scenario-based evaluation of external shocks and policy interventions. Unlike traditional ABMs with static rules, our LLM-powered agents dynamically adapt behaviors to varying conditions, enabling deeper exploration of "what-if" scenarios in urban criminology and policy analysis.

%% file: tables/main_table.tex
\begin{table}[!htbp]
    \small
  \centering
  \caption{Overall performance in simulating spatial crime patterns. 
}
\resizebox{0.8\textwidth}{!}{
    \begin{tabular}{llccccc}
    \toprule
    \textbf{City} & \textbf{Methods} & \multicolumn{1}{l}{\boldmath{}\textbf{HR@1.0 $\uparrow$}\unboldmath{}} & \multicolumn{1}{l}{\boldmath{}\textbf{HR@1.5 $\uparrow$}\unboldmath{}} & \multicolumn{1}{l}{\boldmath{}\textbf{HR@2.0 $\uparrow$}\unboldmath{}} & \multicolumn{1}{l}{\boldmath{}\textbf{JSD $\downarrow$}\unboldmath{}} & \multicolumn{1}{l}{\boldmath{}\textbf{RMSE\tiny{(E-04)} \small 
    $\downarrow$}\unboldmath{}} \\
    \midrule
    \multirow{6}[2]{*}{\textbf{Chicago}} & ABM-Random & 0.2851 & 0.4027 & 0.4796 & 0.2066 & 5.60 \\
       & ABM-Routine~\cite{zhu2021agent} & 0.4208 & 0.5068 & 0.6154 & 0.0916 & 3.50 \\
       & ABM-Hotspot~\cite{weisburd2017can} & 0.4072 & 0.4932 & 0.6109 & \textbf{0.0774} & 2.80 \\
       & ABM-Burglary~\cite{malleson2010crime} & 0.4217 & 0.5542 & 0.6607 & 0.1268 & 4.00 \\
       & DL-UVI~\cite{fan2023urban}  & 0.3057 & 0.4672 & 0.5982 & 0.2521 & 7.00 \\
       & LLM-CrimeMind & \textbf{0.5221} & \textbf{0.6239} & \textbf{0.7157} & 0.0838 & \textbf{1.62} \\
    \midrule
    \multirow{6}[2]{*}{\textbf{Dallas}} & ABM-Random & 0.2465 & 0.3522 & 0.4307 & 0.2180 & 19.5 \\
       & ABM-Routine & 0.2564 & 0.3077 & 0.3675 & 0.2052 & 19.2 \\
       & ABM-Hotspot & 0.3162 & 0.4701 & 0.5556 & 0.2121 & \textbf{18.0} \\
       & ABM-Burglary~ & 0.2735 & 0.3761 & 0.4615 & 0.3427 & 101 \\
       & DL-UVI  & 0.3504 & 0.5128 & 0.5897 & 0.2642 & 19.8 \\
       & CrimeMind & \textbf{0.4749} & \textbf{0.5517} & \textbf{0.6375} & \textbf{0.0887} & 18.5 \\
    \midrule
    \multirow{6}[2]{*}{\textbf{Los Angeles}} & ABM-Random & 0.2629 & 0.3741 & 0.4307 & 0.1298 & 5.88 \\
       & ABM-Routine & 0.2644 & 0.3830 & 0.4681 & 0.0874 & 2.73 \\
       & ABM-Hotspot & 0.2705 & 0.3647 & 0.4590 & \textbf{0.0765} & \textbf{2.22} \\
       & ABM-Burglary & 0.2340 & 0.2340 & 0.2340 & 0.6514 & 72.1 \\
       & DL-UVI & 0.2979 & 0.4134 & 0.4894 & 0.2190 & 2.72 \\
       & CrimeMind & \textbf{0.4053} & \textbf{0.4962} & \textbf{0.5568} & 0.1022 & 2.43 \\
    \midrule
    \multirow{6}[2]{*}{\textbf{New York}} & ABM-Random & 0.2224 & 0.3274 & 0.4258 & 0.4035 & 25.4 \\
       & ABM-Routine & 0.3287 & 0.3986 & 0.5035 & 0.3634 & 22.9 \\
       & ABM-Hotspot& 0.2238 & 0.3287 & 0.4336 & 0.3897 & 24.0 \\
       & ABM-Burglary& 0.2587 & 0.3706 & 0.5035 & 0.2894 & 18.1 \\
       & DL-UVI& 0.3007 & 0.3636 & 0.4126 & 0.7188 & \textbf{12.8} \\
       & CrimeMind & \textbf{0.3776} & \textbf{0.4425} & \textbf{0.5164} & \textbf{0.1750} & 15.3 \\
    \bottomrule
    \end{tabular}%
    }
  \label{tab:main_results}%
\begin{tablenotes}
\centering
\item $\uparrow$ indicates higher is better; $\downarrow$ indicates lower is better.  
\end{tablenotes}

\end{table}%


%% file: 6_conclusion.tex
\section{Conclusion}
\label{sec: conclusion}

We propose \textbf{CrimeMind}, a novel LLM-based framework for simulating spatial crime patterns by combining criminological theory, structured urban data, and unstructured perceptual cues. Our self-alignment pipeline effectively bridges human-perceived safety and agent reasoning, improving decision realism. Experiments show that CrimeMind outperforms traditional agent-based models and deep learning baselines across multiple U.S. cities in both spatial accuracy and hotspot prediction. Ablation studies underscore the importance of theory-informed reasoning, multimodal context, and LLM capabilities. The framework also supports counterfactual simulation and policy evaluation, providing an evaluation tool for safety planning and crime prevention. A limitation lies in our current use of EPR-based (non-LLM) mobility modeling, which we plan to enhance in future work.


%% file: 7_Appendix.tex
\newpage
\appendix
\section{Appendix}   

\subsection{Additional Experiments}

\subsubsection{Ablation Study}   
\label{appendix: ablation}
To further understand the internal mechanisms driving CrimeMind’s strong performance, we conduct a detailed ablation analysis in the Chicago setting. 
We conduct an ablation study by progressively removing core inputs from the CrimeMind framework: (i) Routine Activity Theory-based reasoning, (ii) static urban features, (iii) street-level perceived safety scores, and (iv) semantic street descriptions.

\textbf{Effect of RAT-based Reasoning.} Removing the RAT module leads to a noticeable drop in HR@2.0 (from 0.7157 to 0.6222), despite achieving a little lower JSD. This shows that by incorporating crime theory into the cognitive architecture, llm agent can make more realistic crime decision that produces more accurate crime patterns.

\textbf{Effect of Urban Context Information.} We further ablate three types of urban context: specifically, removing the \textit{safety score} causes a large drop in HR@2.0 and increases both JSD and RMSE, indicating the importance of fine-grained visual perception for crime risk; excluding \textit{semantic descriptions} slightly degrades performance, showing that textual understanding may offer complementary high-level cues; without \textit{static urban features} (e.g., demographics), performance drops in all metrics, showing the role of socio-economic data in modeling vulnerability and routine patterns. If we exclude all the urban context, the model still outperforms traditional baselines in HR@2.0, demonstrating the inherent strength of LLM-based agent reasoning. However, the performance gap to the full model shows that urban context is essential for simulating the real-world crime patterns.

Overall, these results demonstrate that each module contributes to CrimeMind’s success. The combination of decision-theoretic modeling (via RAT), multi-modal urban context, and LLM reasoning provides a comprehensive foundation for accurate spatial crime simulation. 

\input{tables/ablation}

\input{tables/llm}

\subsubsection{Effect of LLM Types}
To investigate the impact of different large language models on crime simulation, we substitute the reasoning module in CrimeMind with several representative LLMs, including variants from the Qwen, LLaMA, and GPT families. As shown in Table~\ref{tab:llm}, all LLMs exhibit competitive performance, demonstrating that CrimeMind maintains robust generalization capabilities across a diverse range of model architectures and scales. 
Within the Qwen2.5 series, the performance generally improves with larger model size, Qwen2.5-32B achieves better HR@2.0 (0.7203) and lower JSD (0.0772) than Qwen2.5-7B. However, Qwen2.5-72B, despite being the largest, slightly underperforms on HR@1.0 and JSD, indicating diminishing returns or increased decision variance at a large scale. 
Among all evaluated models, LLaMA3-70B achieves the best overall performance (HR@1.0 = 0.5605, JSD = 0.0685, RMSE = 0.000080), indicating both strong local reasoning and high distributional alignment. In contrast, GPT-4.1-Nano lags behind other models, with notably higher RMSE and lower hotspot recall, implying weaker alignment with empirical crime patterns.
These findings suggest that while CrimeMind is generally effective across different LLMs, model choice still influences the quality of simulation.

\begin{figure}[htbp]
    \begin{center}
        \includegraphics[width=0.5\linewidth]{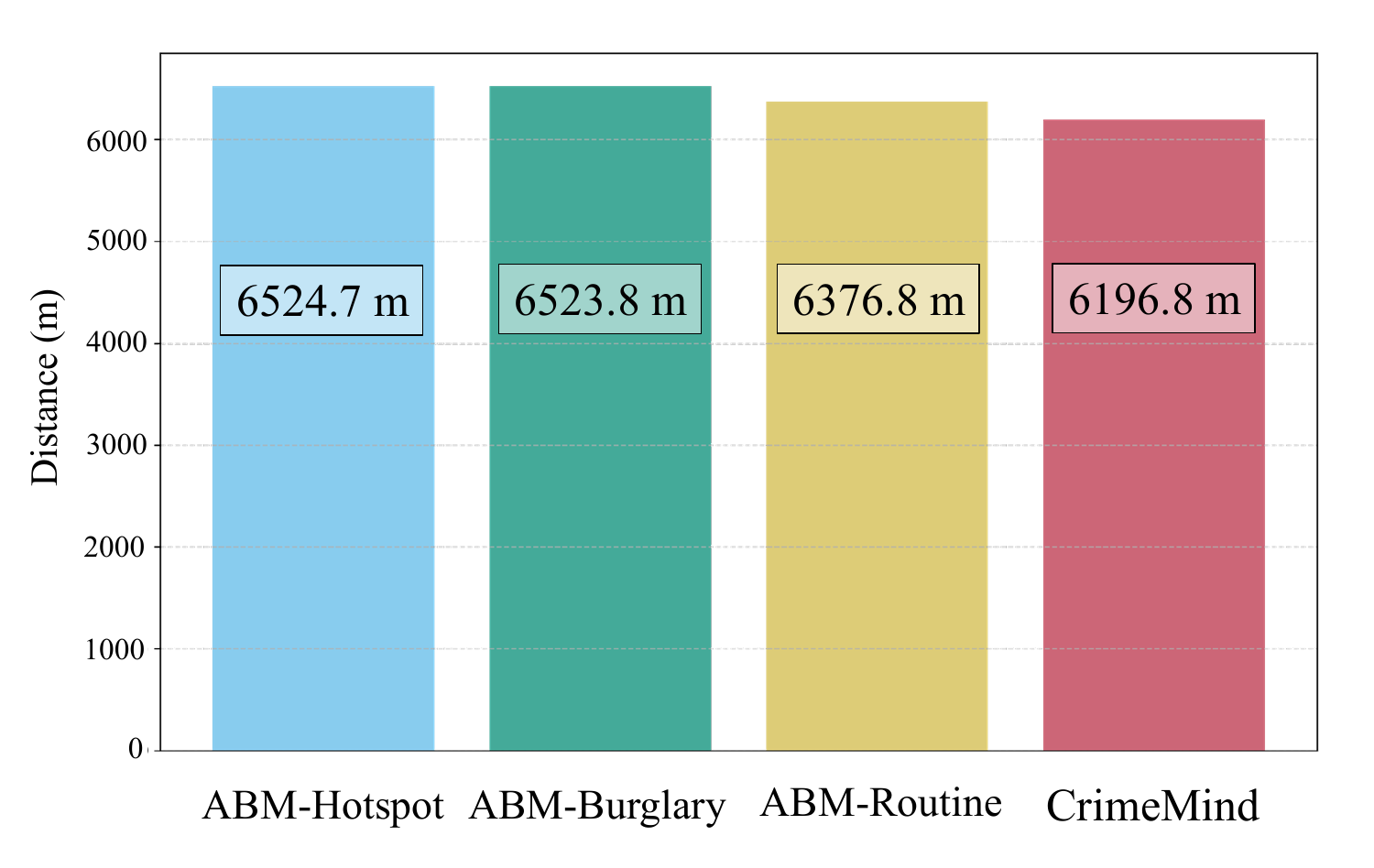}
        \caption{Average distance between crime locations and residences of criminals in Chicago.}   
        \label{Fig:appendix_distance_chicago}
    \end{center}
\end{figure}

\begin{figure}[H]
    \begin{center}
        \includegraphics[width=1.0\linewidth]{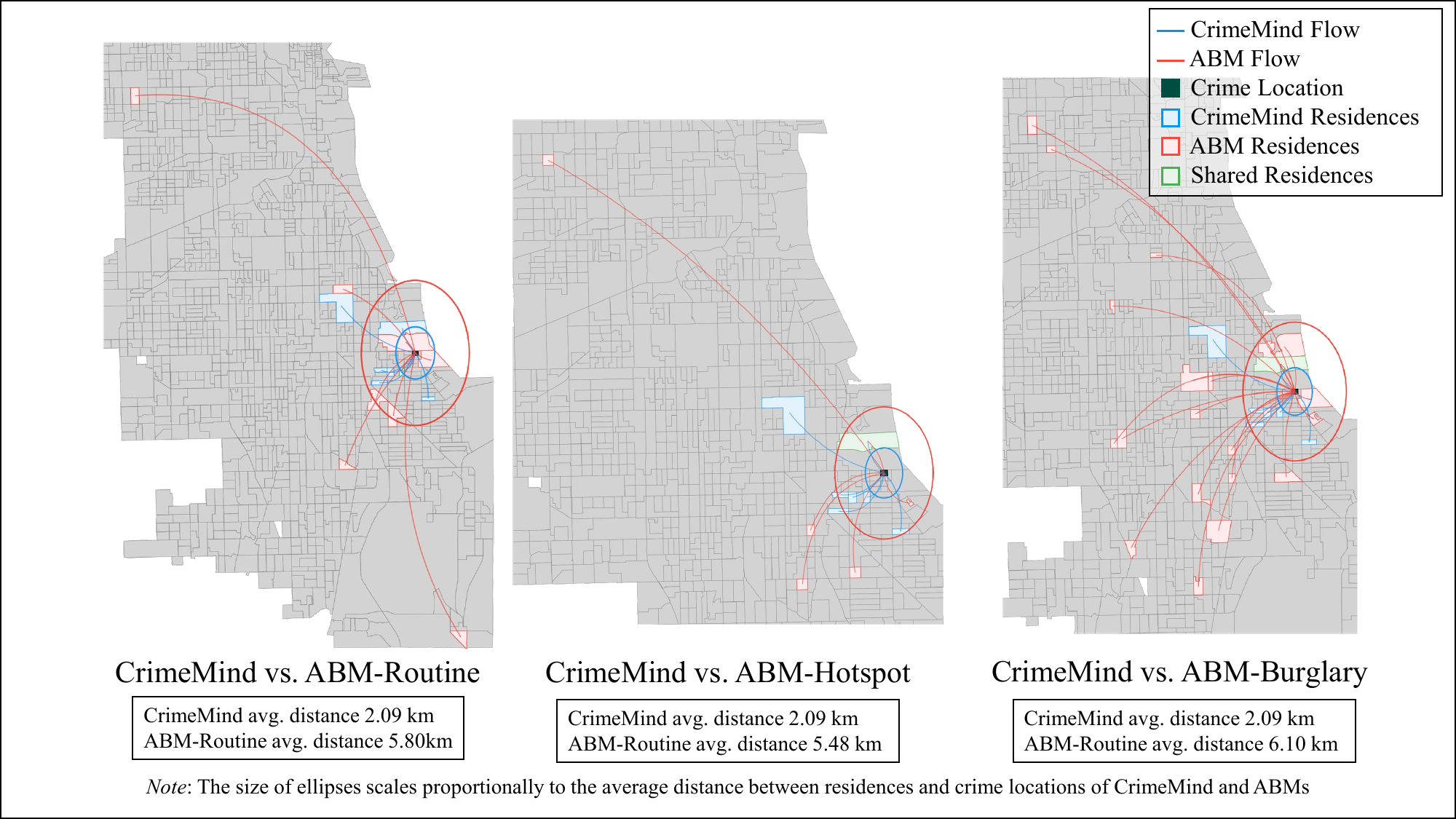}
        \caption{Criminal residences flow in Chicago with different methods for CBG 170318439005.}   
        \label{Fig:appendix_flow_chicago}
    \end{center}
\end{figure}

\subsubsection{Case Study on Residential Distribution of Criminals}

Research in criminology shows that criminals are usually more likely to commit crimes near their residences \cite{curtis-ham-2022, bernasco-2009, massey-1995}. This behavioral pattern may drive away higher socioeconomic status residents and ultimately increase the extent of racial segregation or other socioeconomic segregation \cite{OFLAHERTY2007391, krivo-2009}. As a result, the more severe the segregation in the residential area is, the higher the violent crime rate should be.

\begin{figure}[htbp]
    \begin{center}
        \includegraphics[width=1.0\linewidth]{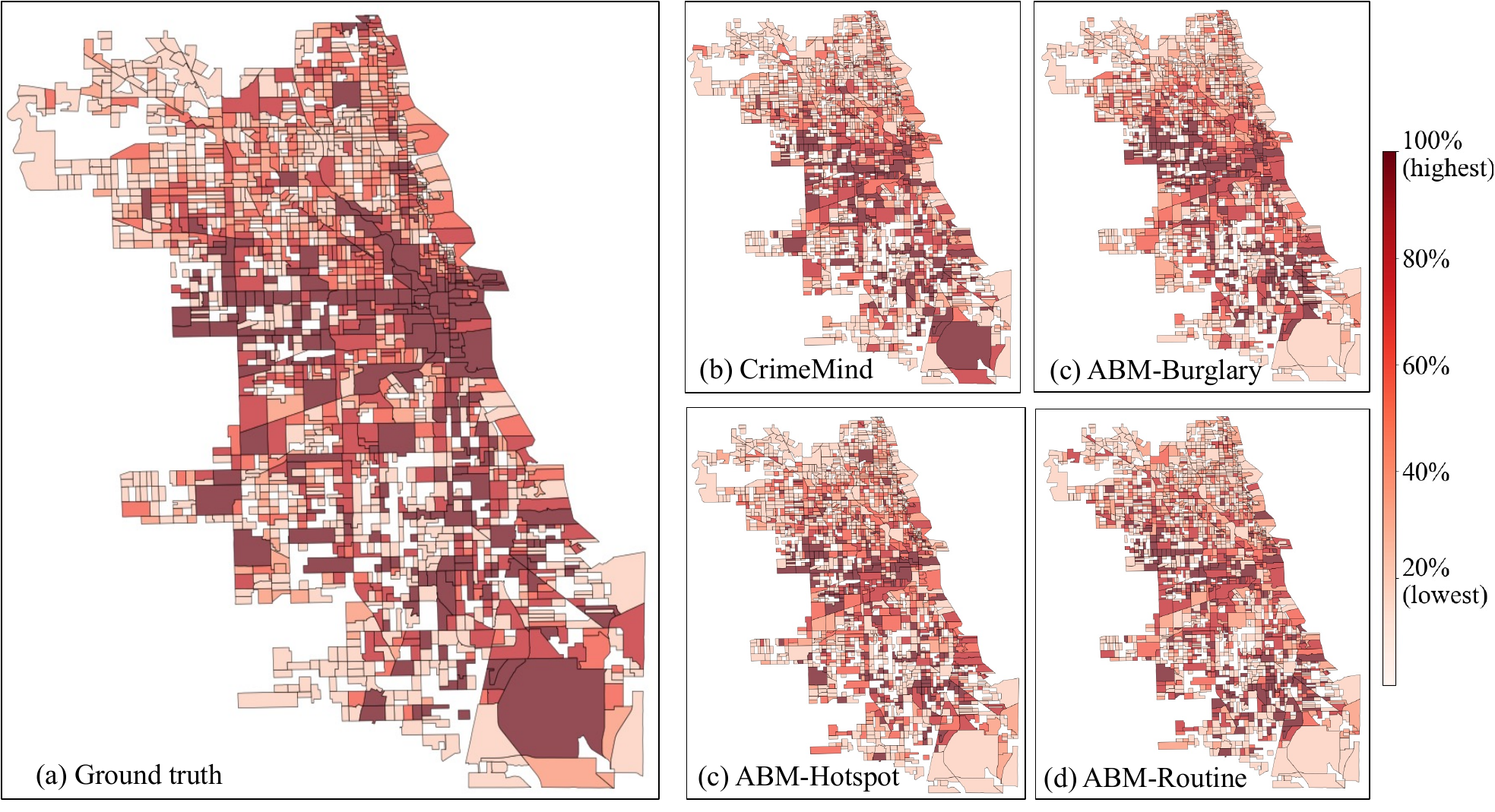}
        \caption{Crime heatmap in Chicago with different methods.} 
        \label{Fig:appendix_heatmap_chicago}
    \end{center}
\end{figure}

Heatmap in Figure \ref{Fig:appendix_heatmap_chicago} and experiment results in Table \ref{tab:main_results} show that CrimeMind is generally more effective in crime prediction than traditional ABMs. An obvious trend is observed as demonstrated in Figure \ref{Fig:appendix_distance_chicago} that traditional ABMs usually have a longer distance between residences and crime locations, which indicates that traditional ABMs fail to support the behavioral patterns in which criminals are more likely to commit crimes near their residences in comparison with CrimeMind. 

We chose a specific CBG in Chicago for the case study as shown in Figure \ref{Fig:appendix_flow_chicago}. In this case, the average distance between crime location and the criminals' residences of all three ABMs is significantly larger than that of CrimeMind. The case shows that the criminals of CrimeMind usually come from the residences closer to the crime location, while the criminals of traditional ABMs usually have more scattered residences and spread throughout the entire map of Chicago, and are also farther apart from the crime location. This reflects that CrimeMind can better display the segregation phenomenon within the urban environment, while achieving more accurate predictions on crime and criminals' trajectories.

\subsection{Supplementary Figures }

\begin{figure}[!hb]
    \begin{center}
        \includegraphics[width=0.8\linewidth]{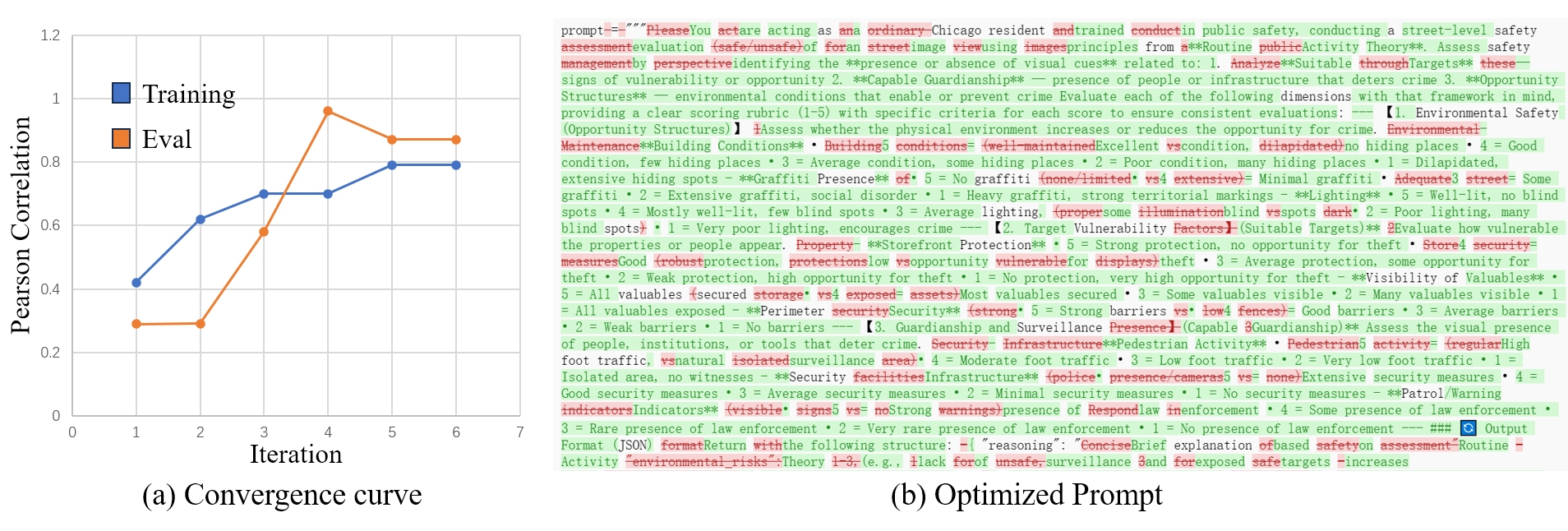}
        \caption{Illustration of the effectiveness of the perceived safety alignment.}   
        \label{Fig:textgrad}
    \end{center}

\end{figure}

\begin{figure}[!ht]
    \begin{center}
        \includegraphics[width=1.0\linewidth]{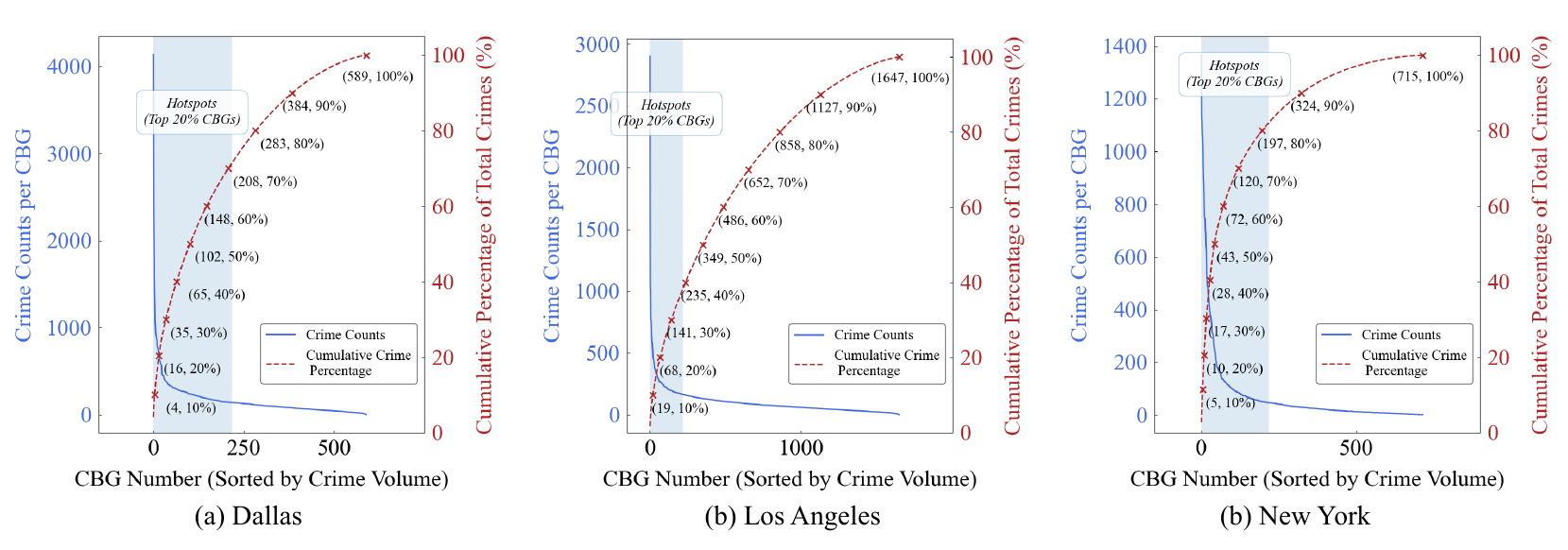}
        \caption{Crime hotspots of each city.}   
        \label{Fig:appendix_hotspots}
    \end{center}
\end{figure}

\begin{figure}[!htbp]
    \begin{center}
        \includegraphics[width=1.0\linewidth]{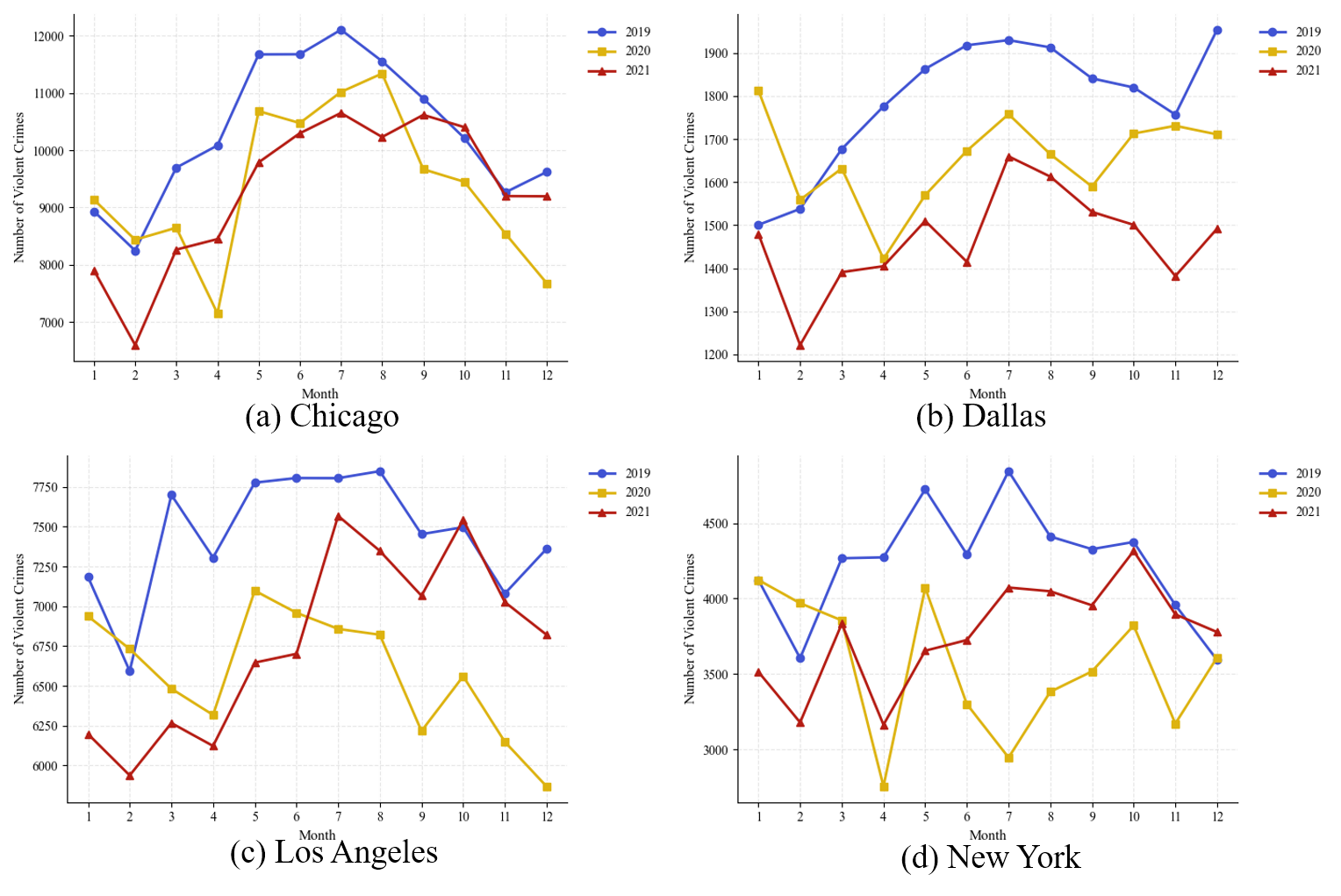}
        \caption{Crime change across time in each city.}   
        \label{Fig:appendix_crimeTime}
    \end{center}
\end{figure}

\begin{figure}[!htbp]
    \begin{center}
        \includegraphics[width=1.0\linewidth]{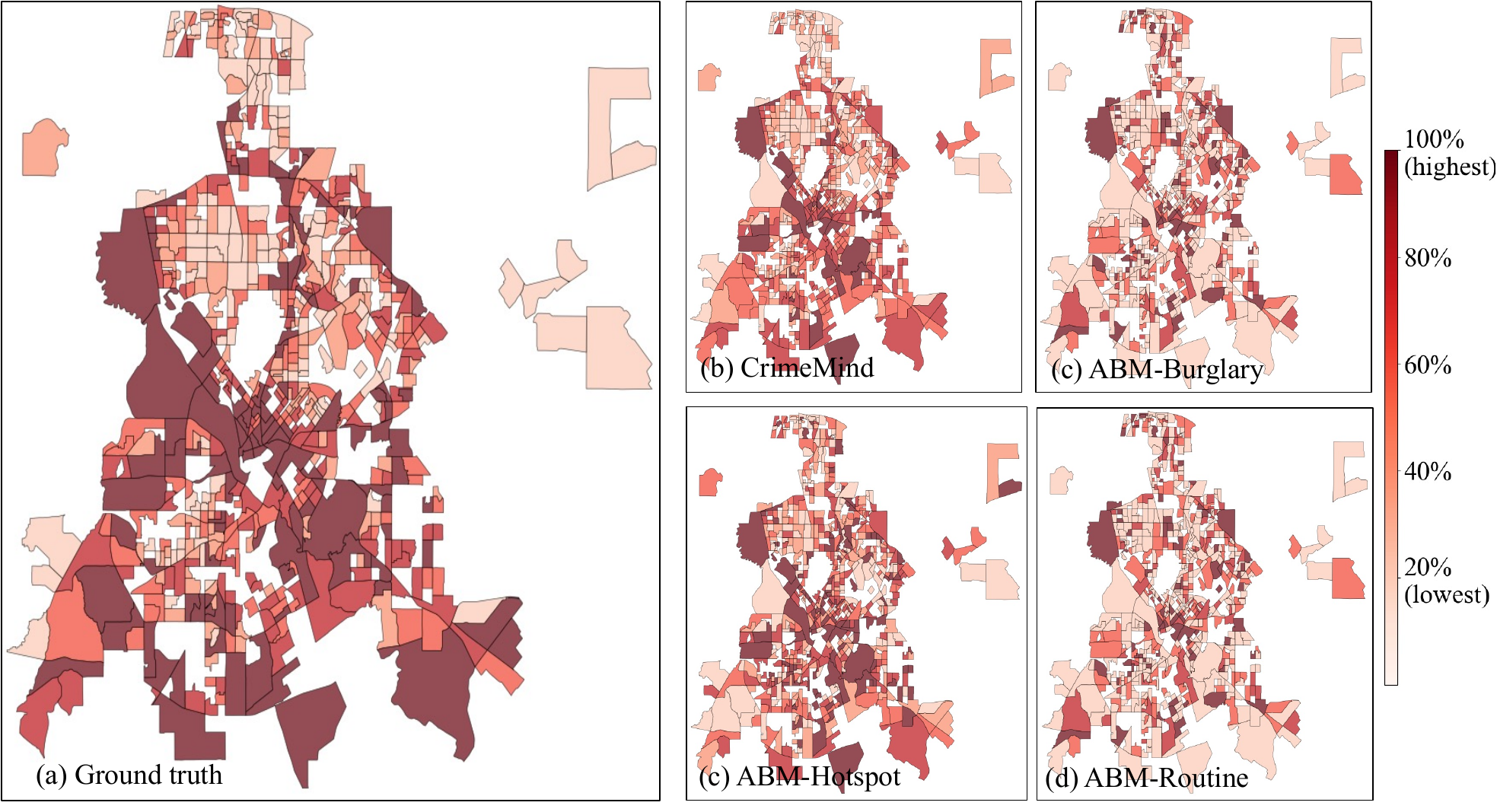}
        \caption{Crime heatmap in Dallas with different methods.}   
        \label{Fig:appendix_heatmap_dallas}
    \end{center}
\end{figure}

\begin{figure}[!htbp]
    \begin{center}
        \includegraphics[width=1.0\linewidth]{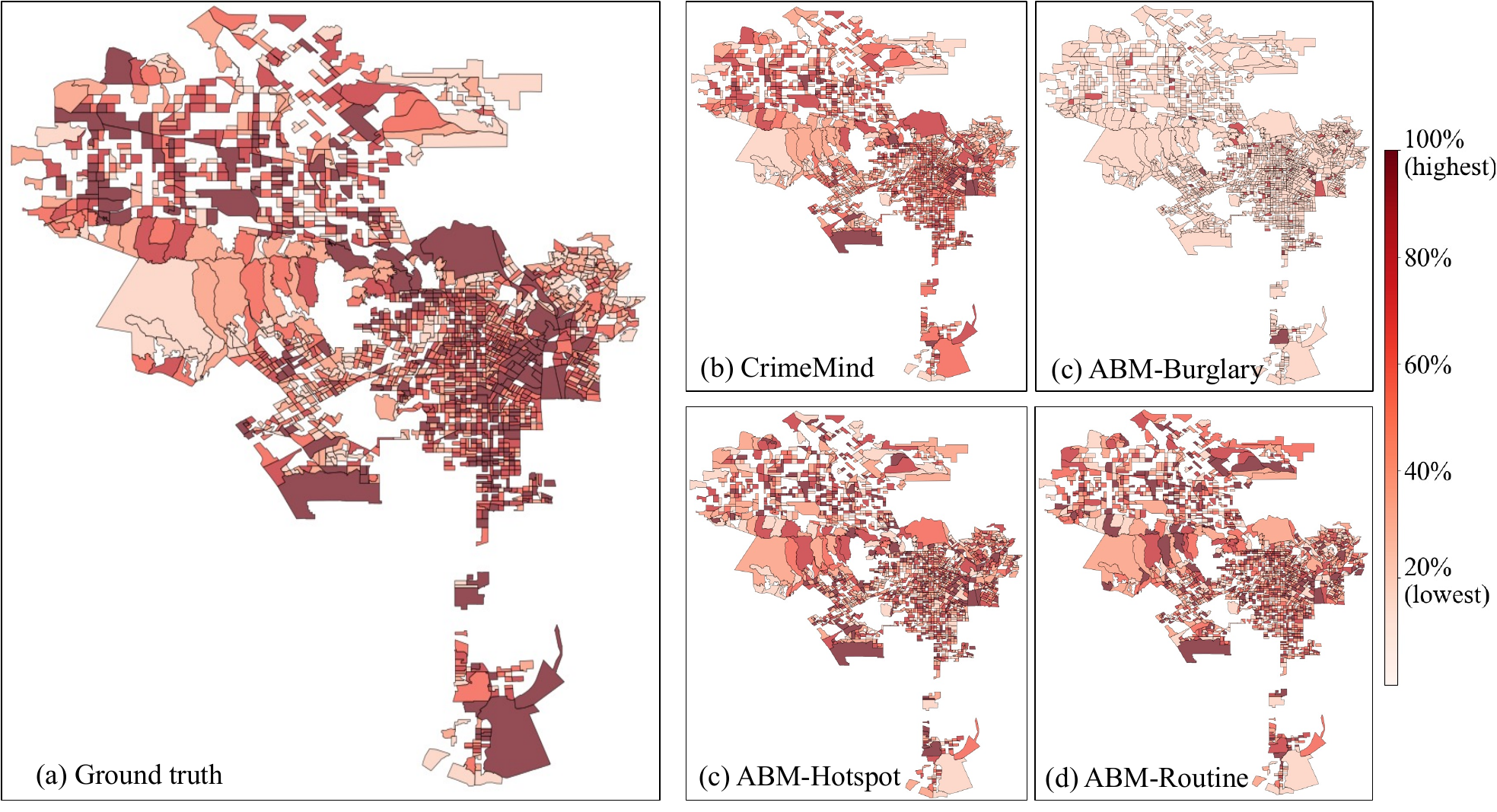}
        \caption{Crime heatmap in Los Angeles with different methods.}   
        \label{Fig:appendix_heatmap_LA}
    \end{center}
\end{figure}

\begin{figure}[!htbp]
    \begin{center}
        \includegraphics[width=1.0\linewidth]{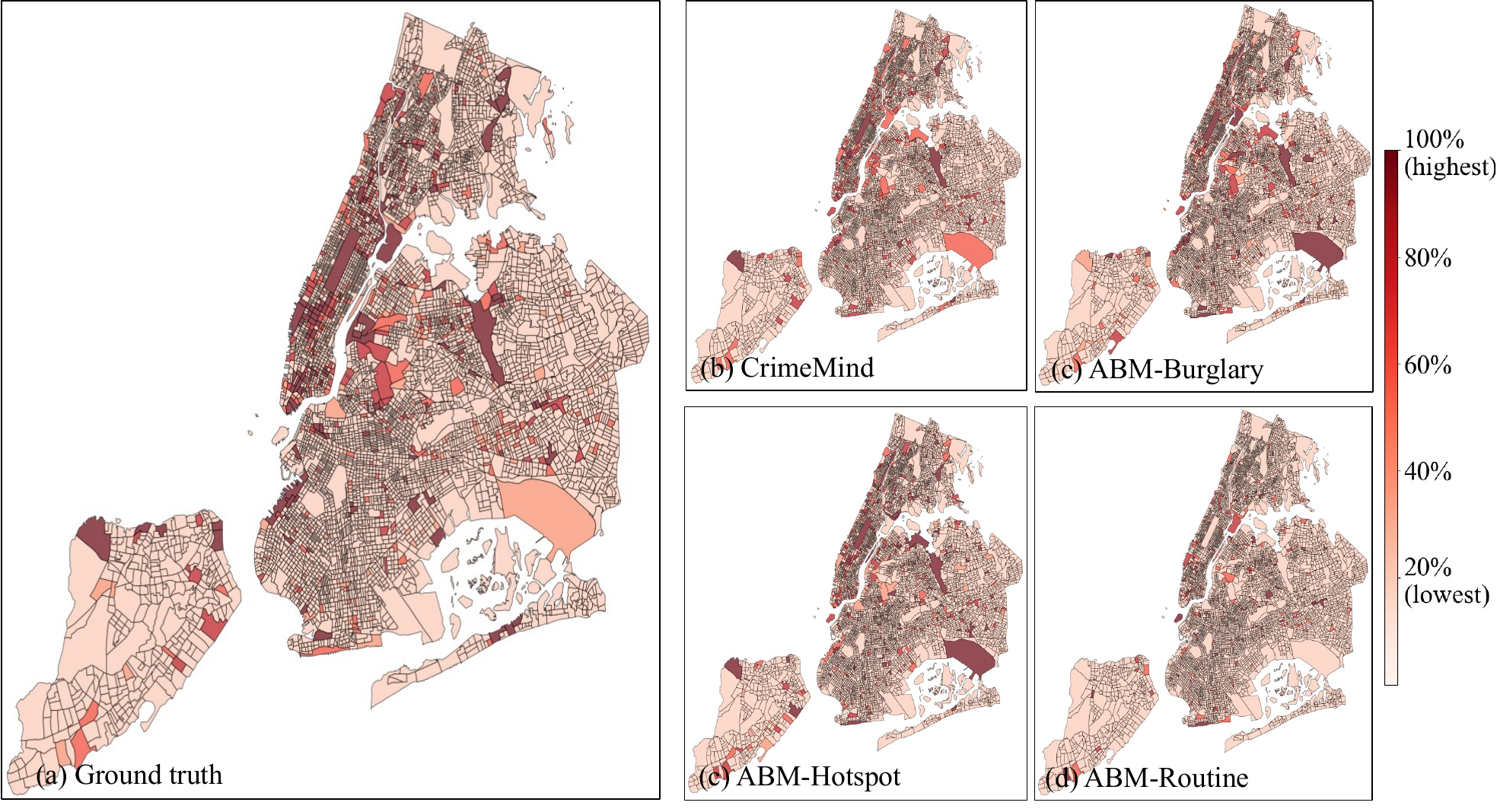}
        \caption{Crime heatmap in New York with different methods.}   
        \label{Fig:appendix_heatmap_NY}
    \end{center}
\end{figure}

\newpage
\subsection{Discussions}
\subsubsection{Limitations}
\label{appdix: limitation}
While CrimeMind presents a novel integration of LLM-based agents and multimodal urban environments for crime simulation, several limitations remain. First, large language models may introduce potential biases inherited from pretraining data, which may affect agents’ decision-making in unintended ways, particularly when modeling sensitive behaviors like crime. To mitigate this, we design prompts that cast the LLM agent as an expert criminologist reasoning about offenders' decisions, which helps reduce direct imitation of unethical behavior. However, this mitigation is imperfect and warrants further exploration, such as through fine-tuning with domain-specific, ethically aligned datasets. Also, due to computational constraints, the current simulation uses a fixed number of agents and discretized time steps. The agent mobility is modeled using a non-LLM-based exploration model (EPR), which may limit the agents' ability to adaptively perceive and respond to environmental cues in a human-like manner. Enhancing agent mobility with more dynamic, LLM-informed navigation strategies, especially in complex urban environments, is an important direction for future work.


\subsubsection{Code of Ethics}
\label{appendix: ethic}
The datasets involved in this work are all open-source, which poses no problem regarding privacy and copyright. Specifically, street view images are obtained via the Google Street View API, and crime data is sourced from official city open data portals. SafeGraph data are aggregated at the CBG level, ensuring no individual-level or personally identifiable information is included. 
We cite the resources in Section~\ref{subsec: dataset}, and Section~\ref{subsec: exp_setup}.

\subsubsection{Broader Impacts}
\label{appendix: impact}
Our method has positive broader impacts. Our method aims to enhance the realism, interpretability, and adaptability of urban crime simulation through LLM-driven agents and multimodal environmental context. This can support city planners, criminologists, and policymakers in evaluating the impacts of socio-political shocks and public safety interventions. By grounding agent decision-making in criminological theory and environmental perception, CrimeMind offers a novel framework for understanding complex human-environment interactions and crime dynamics in urban safety.


\subsubsection{Computational Cost}
\label{appendix: cost}
Our framework relies on querying LLM APIs for agent decision-making, which constitutes the major computational bottleneck. Specifically, each criminal agent's behavior at every time step involves at least one prompt-response interaction with the LLM, leading to a high volume of requests in large-scale simulations. Typically, simulating 5,000 agents over 50 time steps takes about 3 hours to complete (parallelized in batches of 64 with moderate latency). The cost is primarily time-bound, as LLM inference via API services is performed off-device and billed by usage.

\subsection{Prompt Example}
\subsubsection{Prompt for criminal agents}
\input{prompts/criminal}

\subsubsection{Prompt for evaluating the perceived score}
\input{prompts/safety}

%% file: tables/ablation.tex
\begin{table}[htbp]
\small
  \centering
  \caption{Ablation studies. Bold indicates the best performance.}
  \resizebox{0.9\textwidth}{!}{
    \begin{tabular}{lccccc}
    \toprule
    \textbf{Methods} & \multicolumn{1}{l}{\boldmath{}\textbf{HR@1.0 $\uparrow$}\unboldmath{}} & \multicolumn{1}{l}{\boldmath{}\textbf{HR@1.5 $\uparrow$}\unboldmath{}} & \multicolumn{1}{l}{\boldmath{}\textbf{HR@2.0 $\uparrow$}\unboldmath{}} & \multicolumn{1}{l}{\boldmath{}\textbf{JSD $\downarrow$}\unboldmath{}} & \multicolumn{1}{l}{\boldmath{}\textbf{RMSE\tiny{(E-04)} \small 
    $\downarrow$}\unboldmath{}} \\
    \midrule
    CrimeMind  & \textbf{0.5221} & \textbf{0.6239} & \textbf{0.7157} & 0.0838 & \textbf{1.62} \\
     - w/o RAT Module & 0.3644 & 0.5022 & 0.6222 & \textbf{0.0754} & 4.26 \\
     - w/o All Urban Context & 0.3925 & 0.5187 & 0.6495 & 0.1302 & 2.13 \\
     - w/o Safety Score & 0.4652 & 0.5826 & 0.6957 & 0.1603 & 3.41 \\
     - w/o Semantic Descriptions & 0.4820 & 0.5995 & 0.7081 & 0.1286 & 2.93 \\
     - w/o Static Urban Features & 0.4980 & 0.6123 & 0.7102 & 0.0945 & 2.56 \\
    \bottomrule
    \end{tabular}%
    }
  \label{tab:ablation}%
\end{table}%

%% file: tables/llm.tex

\begin{table}[!htbp]
  \centering
  \caption{Comparison of Different LLMs}
   \resizebox{0.75\textwidth}{!}{
    \begin{tabular}{lccccc}
    \toprule
      & \multicolumn{1}{l}{\boldmath{}\textbf{HR@1.0 $\uparrow$}\unboldmath{}} & \multicolumn{1}{l}{\boldmath{}\textbf{HR@1.5 $\uparrow$}\unboldmath{}} & \multicolumn{1}{l}{\boldmath{}\textbf{HR@2.0 $\uparrow$}\unboldmath{}} & \multicolumn{1}{l}{\boldmath{}\textbf{JSD $\downarrow$}\unboldmath{}} & \multicolumn{1}{l}{\boldmath{}\textbf{RMSE\tiny{(E-04)} \small 
    $\downarrow$}\unboldmath{}} \\
    \midrule
    Qwen2.5-7B & 0.5221 & 0.6239 & 0.7157 & 0.0838 & 1.62 \\
    Qwen2.5-32B & 0.5410 & 0.6380 & 0.7203 & 0.0772 & 1.48 \\
    Qwen2.5-72B & 0.5182 & 0.6255 & 0.7189 & 0.0815 & 0.98 \\
    Qwen3-8B & 0.5570 & 0.6490 & \textbf{0.7285} & 0.0691 & 1.32 \\
    LLaMA3-8B & 0.5406 & 0.6385 & 0.7021 & 0.0746 & 1.42 \\
    LLaMA3-70B & \textbf{0.5605} & \textbf{0.6532} & 0.7223 & \textbf{0.0685} & \textbf{0.80} \\
    GPT4.1-Nano & 0.4589 & 0.5642 & 0.6588 & 0.0943 & 2.24 \\
    GPT4o-mini & 0.5436 & 0.6178 & 0.7102 & 0.0782 & 1.57 \\
    \bottomrule
    \end{tabular}%
    }
  \label{tab:llm}%
\end{table}

%% file: prompts/criminal.tex
\textbf{System prompt}:
\begin{Verbatim}[fontsize=\tiny, frame=single, samepage=false, breaklines=true, breakanywhere=true] 
    You are an expert criminologist specializing in predicting the decision-making processes of potential criminal agents. Your task is to simulate a detailed internal reasoning process for a given criminal agent(an agent very likely to commit a crime) and context, determining if the agent will commit a crime. Your response should be ONLY the JSON object with no additional text preceding or following it.
    If a crime is predicted, output:
    {  "status": true,
        "objective_id": string, (the agent_id of the target)
        "reasoning": string
    }
    If no crime is predicted, output:
    {
        "status": false,
        "reasoning": string
    }
    "reasoning" should be a detailed explanation of the decision-making process, including the factors considered and their implications for the agent's likelihood to commit a crime.Remember to output ONLY the JSON with no additional text.
\end{Verbatim}

\textbf{Prompt}: 
\begin{Verbatim}[fontsize=\tiny, frame=single, samepage=false, breaklines=true, breakanywhere=true] 
    I want you to think step-by-step through the agent's potential decision-making process. Consider the following instructions carefully:
    **Overall Contextual Information:**
    The agent is in {city}, the mayor is {mayor}, and the party is {party}. This political and administrative context might influence the city's general approach to law enforcement and social programs, potentially captured by the overarching strategy: {strategy}.

    **Step 1: Analyze Agent's Motivation (Likelihood of Intent)**
    Based on the agent's attributes, evaluate their potential motivation to commit a crime."""+f"""
        - Agent ID: {criminal['agent_id']}
        - Gender: {criminal['gender']}
        - Race: {criminal['race']}
        - Residence: {criminal['residence']}
        - Historical Trajectory: {criminal['historical_trajectory']} (What does this suggest about their patterns or desensitization to crime?)
        - Criminal Records: {criminal['criminal_record']} (How extensive and recent is this record? Does it show a pattern of specific crimes? Does it indicate desperation or opportunism?)
        - Current Location: {criminal['current_location']} (Is this location familiar or advantageous for criminal activity?)
    *Self-Correction/Refinement for Motivation:* Are there any conflicting indicators in the agent's profile? For example, a long record but a recent positive trajectory, or vice versa. How does the agent's profile align with common criminological theories of motivation (e.g., strain theory, social learning theory, rational choice theory)?

    **Step 2: Evaluate Target Suitability (Opportunity Assessment - Targets)**
    Consider the provided potential targets and how they might appear to this specific agent.
        - Potential Targets: {target_str} (Describe the nature of these targets. Are they high-value, low-risk, opportunistic, etc.? Which, if any, would be most appealing to an agent with the profile from Step 1?)
    *Self-Correction/Refinement for Target Suitability:* Would all targets be equally appealing, or does the agent's historical trajectory or criminal record suggest a preference for certain types of targets? How does the agent's current location relate to these potential targets in terms of accessibility?

    **Step 3: Evaluate Absence of Guardianship (Opportunity Assessment - Environment & Deterrence)**
    Assess the perceived risk and ease of committing a crime in the current environment.
        - City Description & Environmental Safety Score: (Implicitly, consider the overall safety and typical guardianship levels based on the provided CBG attributes).
        - Police Presence: Number of Patrolling Police Officers: {police_count} (Is this number high or low for the area? How visible and active are they likely to be?)
        - CBG (Census Block Group) Attributes:   
            - Current CBG Description: {desc}
            - Environmental Safety Score(0-1): {score}.Environmental safety score is a measure of the safety and security of the area, with 0 being very unsafe & lack of guardianship & suitable for crime, and 1 being very safe & fully under guardianship & unsuitable for crime. Low environmental safety score may encourage crime motivation. 
            - Number of POIs: {len(cbg['poi'])} (Do these POIs increase natural surveillance or offer more potential targets/cover?)
            - Total population: {population} (Does this population density increase anonymity or surveillance?)
            - average\_income($): {income} (Does this suggest wealth disparity that might motivate crime, or resources for better security?)
            - poverty\_ratio: {poverty_ratio} (Is there economic desperation that might lower the threshold for criminal behavior?)
            - housing\_value ($):{housing_value} (Does this indicate affluent targets or well-protected areas?)
    *Self-Correction/Refinement for Absence of Guardianship:* Are there any CBG attributes that seem contradictory (e.g., high income but also high crime rate)? How might the agent interpret these signals? How does the strategy related to the city's administration tie into the perceived level of guardianship or deterrence?
    **Step 4: Synthesize and Make a Prediction**
    Now, weigh the findings from Step 1 (Motivation), Step 2 (Target Suitability), and Step 3 (Absence of Guardianship).
    - How strong is the agent's motivation?
    - Are there highly suitable and accessible targets?
    - Is the perceived risk of getting caught low enough?
    - Consider the overall context: {city}, mayor {mayor}, party {party}, and strategy {strategy}. Does this overarching context sway the decision?
    Based on this synthesis, will the agent likely commit a crime? If yes, which target (`objective\_id`) is the most probable choice given the analysis?
    **Step 5: Formulate JSON Output**
    Based on your detailed step-by-step reasoning above, construct the JSON output.
\end{Verbatim}

%% file: prompts/safety.tex
\begin{Verbatim}[fontsize=\tiny, frame=single, samepage=false, breaklines=true, breakanywhere=true] 
You are acting as a Chicago resident trained in public safety, conducting a street-level safety evaluation of an image using principles from **Routine Activity Theory**. Assess safety by identifying the **presence or absence of visual cues** related to: 1. **Suitable Targets** — signs of vulnerability or opportunity 2. **Capable Guardianship** — presence of people or infrastructure that deters crime 3. **Opportunity Structures** — environmental conditions that enable or prevent crime Evaluate each of the following dimensions with that framework in mind, providing a clear scoring rubric (1-5) with specific criteria for each score to ensure consistent evaluations: --- [1. Environmental Safety (Opportunity Structures)] Assess whether the physical environment increases or reduces the opportunity for crime. - **Building Conditions** • 5 = Excellent condition, no hiding places • 4 = Good condition, few hiding places • 3 = Average condition, some hiding places • 2 = Poor condition, many hiding places • 1 = Dilapidated, extensive hiding spots - **Graffiti Presence** • 5 = No graffiti • 4 = Minimal graffiti • 3 = Some graffiti • 2 = Extensive graffiti, social disorder • 1 = Heavy graffiti, strong territorial markings - **Lighting** • 5 = Well-lit, no blind spots • 4 = Mostly well-lit, few blind spots • 3 = Average lighting, some blind spots • 2 = Poor lighting, many blind spots • 1 = Very poor lighting, encourages crime --- [2. Target Vulnerability (Suitable Targets)** Evaluate how vulnerable the properties or people appear. - **Storefront Protection** • 5 = Strong protection, no opportunity for theft • 4 = Good protection, low opportunity for theft • 3 = Average protection, some opportunity for theft • 2 = Weak protection, high opportunity for theft • 1 = No protection, very high opportunity for theft - **Visibility of Valuables** • 5 = All valuables secured • 4 = Most valuables secured • 3 = Some valuables visible • 2 = Many valuables visible • 1 = All valuables exposed - **Perimeter Security** • 5 = Strong barriers • 4 = Good barriers • 3 = Average barriers • 2 = Weak barriers • 1 = No barriers --- [3. Guardianship and Surveillance (Capable Guardianship)** Assess the visual presence of people, institutions, or tools that deter crime. - **Pedestrian Activity** • 5 = High foot traffic, natural surveillance • 4 = Moderate foot traffic • 3 = Low foot traffic • 2 = Very low foot traffic • 1 = Isolated area, no witnesses - **Security Infrastructure** • 5 = Extensive security measures • 4 = Good security measures • 3 = Average security measures • 2 = Minimal security measures • 1 = No security measures - **Patrol/Warning Indicators** • 5 = Strong presence of law enforcement • 4 = Some presence of law enforcement • 3 = Rare presence of law enforcement • 2 = Very rare presence of law enforcement • 1 = No presence of law enforcement --- ### Output Format (JSON) Return the following structure: { "reasoning": "Brief explanation based on Routine Activity Theory (e.g., lack of surveillance and exposed targets increases risk)", "environmental_safety_score": 1-5, "vulnerability_safety_score": 1-5, "surveillance_safety_score": 1-5, "overall_safety_score": 1-5 } Only use what is clearly visible in the image. If a feature is not visible (e.g., police presence), assume it is absent.
\end{Verbatim}